# Indonesian Earthquake Decision Support System

Spits Warnars H.L.H
Faculty of Information Technology
University of Budi Luhur
Petukangan utara, Jakarta Selatan 12260
INDONESIA
spits@bl.ac.id http://www.spits.8k.com

*Abstract* : Earthquake DSS is an information technology environment which can be used by government to sharpen, make faster and better the earthquake mitigation decision. Earthquake DSS can be delivered as E-government which is not only for government itself but in order to guarantee each citizen's rights for education, training and information about earthquake and how to overcome the earthquake. Knowledge can be managed for future use and would become mining by saving and maintain all the data and information about earthquake and earthquake mitigation in Indonesia. Using Web technology will enhance global access and easy to use. Datawarehouse as unNormalized database for multidimensional analysis will speed the query process and increase reports variation. Link with other Disaster DSS in one national disaster DSS, link with other government information system and international will enhance the knowledge and sharpen the reports.

## 1. Introduction

Monday, May 12$^{th}$,2008 in the middle of the day, around 2:28 p.m. local time, eastern China was hit by 7.9 magnitude quake. More than 40000 people have died so far from the quake centered in Sichuan province. This horrible incident remember us in Indonesia about Aceh's quake on Dec 26$^{Th}$ ,2004 where 9.1 on the Richter scale earthquake just off the west coast of the island of Sumatra, and the tsunami that followed, killed at least 230,000 people in 12 countries, including about 168,000 in Indonesia alone. Aceh's quake was listed as the 12 of the most destructive earthquakes [3]. Possibility Sichuan's quake will be listed for number 13 as the second list for China after Tangshan's quake which had been listed as 7.5 on the Richter scale on July 27$^{th}$, 1976.

After Aceh's quake many earthquakes have struck Indonesia alternately and even other disasters have been a threat for every citizen in this country. Actually an everyday occurrence on earth and more than 3 million earthquakes occur every year, about 8,000 a day, or one every 11 seconds [13] in Indonesia there are 5 to 30 quakes prediction everyday [10]. Government's responsibility to protect the citizen has been done by making National body of disaster management [4][5]. Preparing, saving and distribution logistic become National body of disaster management's responsibility to build information management [11; article 18,26]. Many law's products have been produced as a government's responsibility to give secure life for the citizen. We can not prevent them totally, we have to learn to live with them and need to be prepared all the time, need to learn how to mitigate risk of losses in such events by managing crisis and emergencies correctly[16]. After disaster happens respond must be rapidly and at an optimal level to save lives and help to victims.

## 2. Problem identification

Government have responsibility to declare disaster management start from pre disaster, on disaster and after disaster [2][4]. Every earthquake will create the risks for both life and property and become government's duty for to do the mitigation to fulfill citizen's rights. Learn from old disaster, record it, learn and plan for future mitigation and hope will reduce the disaster risk in the future.

Government as the decision maker which make status and disaster level [4][11] never have the tools which can help them to make faster and better decision with the right and acceptable information. Our government can't be able to make faster and better mitigation decision and the victims will increase by delaying the mitigation decision. If the Government gets the information as an input for their decision, is it the right and acceptable information? Is it only just information which can make happy the decision maker? How about old data, past knowledge, or past earthquake? Do they get information about past earthquakes? How they made the information? How about the failures and successes from past mitigation earthquakes? Do we learn from our past things and become well better and sharpen to make the decisions?

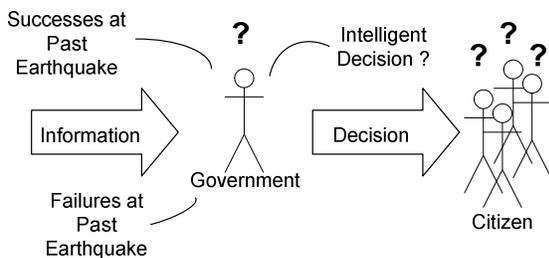

Fig 1. Government when make the earthquake mitigation decision

Do citizen know all earthquakes things from one corridor or they just read from the newspaper? Look at National body of disaster management at www.bakornaspb.go.id, is it live or dead website? The website is supported with forum as a communication for citizen, but if take a look at that website forum, then there only one post which told about desperate hope to make the website become better for citizen and international. Then what's wrong with it? Have National body of disaster management run their obligation to fulfill accountability responsibility to public [11, article 45], guarantee for each citizen's rights for the security, education, training and information about earthquake and how to overcome the earthquake disaster [4; article 26]?

What is going on with this country? Where is the knowledge? Do the knowledge only keep in some public services' head? What is going on if they died? Can this nations' descendant survive for the earthquake problem in the future and how many victims will be recorded? Is it God's anger and does not care about human's life or as human we too stupid to use our head to make better life for this country? Are there some greedy citizens who always keep everything in their hand, never tolerant with transparency? We know we can't run away from the disaster but God Almighty has given us the brain which we can use to serve Him and neighbor.

Information technology must be used and ahead by government to guarantee the citizen's rights to be serviced by government and not only pay for their tax. E-government must be developed to deliver information and public services to citizen in order to be more transparent [7]. Government to citizen (G2C), Government to business (G2B) and government to government (G2G) can be developed as the implementation of E-government [7]. There must be a willing from government to build the system which can help them, their citizen, the next generation, with transparent and accountability. Do not let the knowledge will be buried in some public services' heads and not share to all citizen. Citizen can access anytime, in anywhere and even international can access about earthquake in Indonesia completely. Moreover in the right hand the data and information can be a secret weapon and there will be mining there!

DSS is an approach which can be delivered to make faster and better decision [16]. DSS can save past data, manage it and use it as a part to make the decision and in the future intelligent DSS can be developed. DSS is not only for the decision maker, but for citizen, scholars, researches and people around the world can access for right and acceptable information.

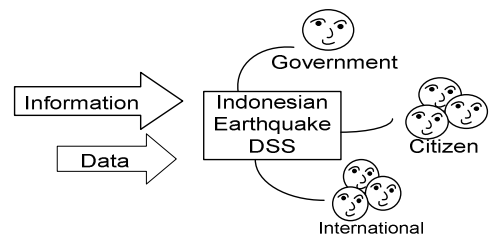

Fig 2. Earthquake DSS for every stakeholder

The DSS as E-government implementation can make comfortable for the citizen about their government's attention for protection upon their live and business. People can access what standard procedure must be declared and done by government to protect their live and business. The standard procedure will distinguished in many different levels disaster's situation and different type of disasters. By clear standard procedure the citizen will quiet in their daily live and business and there is no fluctuation which can disturb the national economy when the disasters have come. The national economy will recover in short time or even will not be influenced when the disasters have come. For the real economic sector, the right handle for the disturbance will have no effect. The bad issue in shortly will bring the fluctuation but bad issue will have no effect if there are some good information.

## 3. Earthquake DSS architecture

Earthquake DSS is an approach in which combine software, hardware and technology in order to deliver the information as support for decision making. Earthquake DSS as an E-government application will be developed with web technology. E-government will make the system which can be accessed from anywhere and anytime in transparency and accountability. Earthquake DSS application can be built with:
1) Client programming like HTML, Javascript or Jscript, Code Style Sheet, and Java Applet.
2) Server Programming like ASP, PHP, JSP, coldfusion or CGI
3) Database can use such as MySQL, Oracle, SqlServer, or postgrace.

The choosing for server programming and database technology will depend on agreement, optional to choose between proprietary or open source. For web hosting service will appoint an approved ISP (Internet Service Provide).

For delivering the information multimedia is an important thing, all the multimedia elements like text, sound, picture, graphic art, animation and video must be included in it. Application and database will be divided into 2 servers for performance and for security there will be needed firewall to control every access. Specifically for database server will have 2 kind of databases are Datawarehouse as UnNormal Structure database for multidimensional analysis, save all information, past data, archive, external data as hypercubes [6]. Transactional is database support for earthquake DSS transaction.

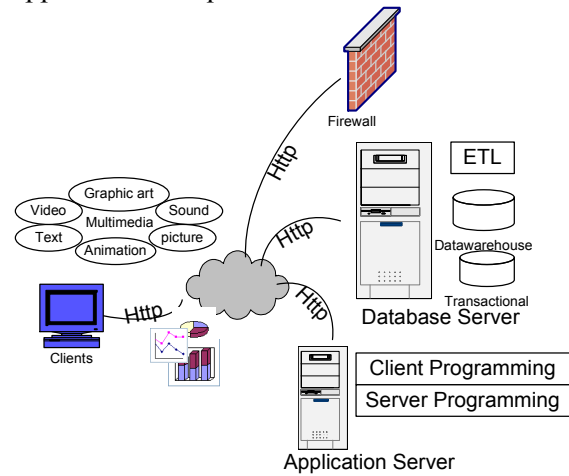

Fig 3. Earthquake DSS architecture

Reports on clients will be delivered both text and chart and can be viewed with many variant and multi dimensional. Drill-down, Roll up, slice and dice is way to show multidimensional in hypercubes. ETL (Extraction, Transformation and Loading) application can be put in database server which automatically will extract data from transactional and external database into datawarehouse. For performance and not disturb the process then data will be captured in Deferred data extraction base on Date and time stamp [6]. For datawarehouse dimensional model will be used snowflake schema. Kinds of user like miners, explorers, farmers, tourists and operators may access the datawarehouse [6].

Datawarehouse earthquake DSS record the disaster location, quantity of victim, infrastructure and property damage [11, article 22]. The Quake table is a fact table which record date, time, longitudinal, latitude, magnitude, epicenter, length, and area of quakes. Other tables is dimensional tables. Quake table will have connection with Stasiun Quake table as table list of earthquake station[10]. To make easy to record area of earthquake then quake table will be connected to regency and provinces tables. In order to record number of dead people then table dead will be connected and also for number of injured people will be recorded in injured table. Where both of dead and injured tables will be connected to medic table as list of medical team who as person in charge for the dead and injured. All the tables dead, injured and medic will be connected to people table as citizen database.

Table people as an improvement to make it easy for citizen's recording. It is time for Indonesia for having the database which can imply per citizen. One identity for one citizen. When a baby is born from Indonesia citizen parents, will get one identity. This single identity will make easy the government to control their people, easy to control the bad people, terrorist and there will be no more people have more than one identity card. Easy for government to recognize their citizen, where and where they was born? when and where they died? what level of their education? is it worthy to get poverty grants? Is it worthy to go for the house of representative? Who has double vote for election? Many things can be done, but the trigger is depend on the government itself.

The building table as record table for every damages from infrastructure, house, office, school, hospital, public places, and all other properties. Table Quake, Dead, Injured and Building are the main datawarehouse for earthquake DSS and other tables as external table from other government institution. Table Stasiun Quake from BMG information system, table regency and province from National administration Information system. Table people as citizen table from national demography information system and table medic from national medical information system. As a conclusion earthquake DSS will be connected with other government information system. The implementation will more difficult but orderliness will be needed for transparency and accountability.

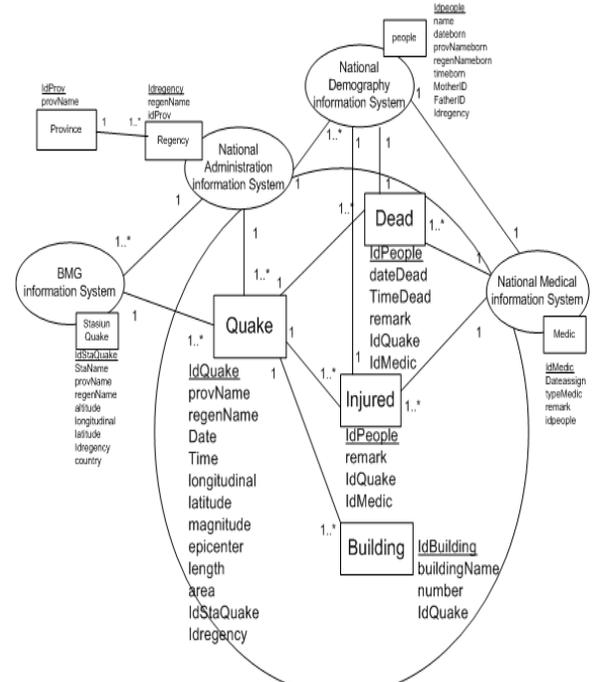

Fig 5. Datawarehouse and connectivity with other internal database

Datawarehouse earthquake DSS would become data mart if there are datawarehouses in other governments information system and of course for harmony will be good if every government institution will having datawarehouse and run their E-government as giving honor and services to the citizen. Earthquake DSS will be more powerful if there are other disaster DSS and as a part of National

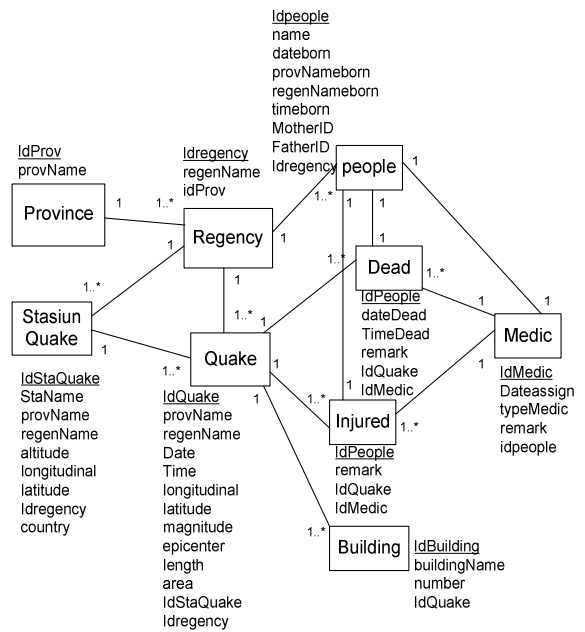

Fig 4. Design entity class diagram datawarehouse

disaster management. Thus other disaster DSS will be researched like :
1) Tsunami disaster DSS.
2) Technology failure disaster DSS.
3) Forest fire disaster DSS.
4) Hurricane disaster DSS.
5) Eruption disaster DSS.
6) Dryness disaster DSS.
7) Landslide disaster DSS.
8) Wave disaster DSS.
9) Flood disaster DSS.
10) Epidemic disaster DSS.

This National Disaster management will have connectivity with other government institution and with international or other governments.

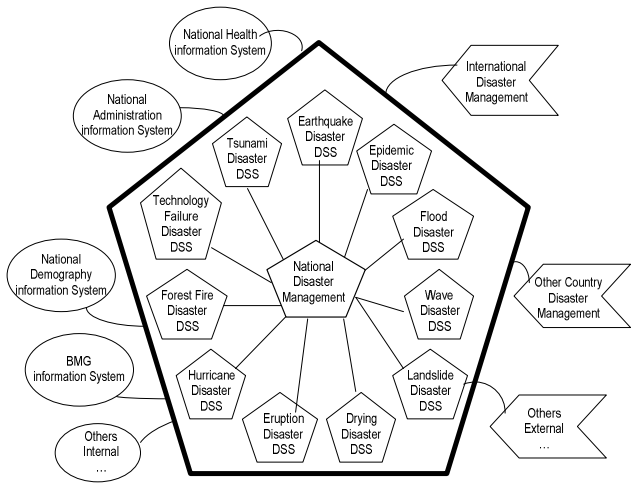

Fig 6. National disaster management and connection with internal and external environment

### 4. Earthquake DSS Scenario

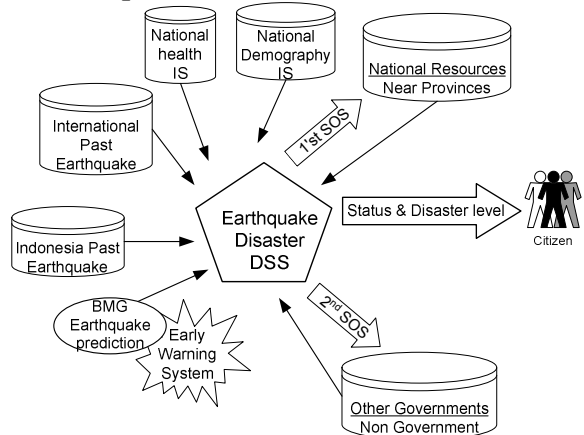

Fig 7. Earthquake DSS scenario

How the earthquake DSS will be run? The scenario below will explain figure 7 about how the earthquake DSS will be run.

1. Earthquake DSS will get an alert from BMG earthquake prediction with its early warning system. An alert will be given by BMG for high risk earthquake which has high magnitude or will hit important places or over population places.

2. Base on BMG earthquake prediction, earthquake DSS will get the information from Past earthquake both in Indonesia or international and make the comparison. Base on the information and the comparison then earthquake DSS will give status and disaster level prediction. Earthquake DSS will predict some area like :
   a. How many national medics we need?
   b. How many international medics we need?
   c. How many died will be?
   d. How many injured will be?
   e. How many national volunteers we need?
   f. How many international volunteers we need?
   g. How many tents we must build?
   h. How many sanitations we must build?
   i. How many food shelters we must build?
   j. How many kilogram rice we must provide?
   k. How many baby feed we must provide?
   l. How many blankets we must provide?
   m. Which locations will be appropriate for refugee's places?
   n. Total lost predictions
   o. Many public places' damages.
   p. And many questions.

   The most needed which must be given attention [11; article 52]
   - Food
   - Clothing
   - sWater
   - Sanitation
   - Rescue Team
   - Health services
   - Psychological services

3. Base on BMG earthquake prediction, earthquake DSS will get the information

about amount of citizens in earthquake area's prediction from National demography Information system.
Formula:
Ac= Amount of citizen at earthquake area.

4. Base on BMG earthquake prediction, earthquake DSS will get the information about amount of medic in earthquake area's prediction from National health Information system. Also from National health Information system will be gotten national standard for maximum handling citizen per medic in disaster time.
   The medical team needed will be gotten from Amount of citizen at earthquake area (Ac) is divided with National standard for maximum handling citizen per medic in disaster time (Nmc).
   Formula:
   Am = Amount of medic at earthquake area.
   Nmc = National standard for maximum handling citizen per medic in disaster time.
   Mn = Medical team needed.
   Mn = Ac / Nmc.
5. If Medical team needed (Mn) more than amount of medic at earthquake area (Am) then lack for medical team will be yielded by subtract Medical team needed (Mn) with Amount of medic at earthquake area (Am).
   Formula:
   Ml = Medical team lack.
   If (Mn>Am) Ml=Mn – Am.
6. In other word if lack for medical team (Ml) more than zero then earthquake DSS will send $1^{st}$ SOS to National resources or near province which can support medical team. The amount medical team which can supply in disaster time from each province will be standardized in National health Information system. In disaster time the area which has been hit by disaster can propose help to near province area. [11, article 28,29,30,31].In disaster time the head of National body of disaster management propose all the resources [11, article 27,28].
7. After get the amount of support medical team from near provinces and national resources then if still there is medical team lack, earthquake DSS will send $2^{nd}$ SOS to international. In disaster time National body of disaster management can propose help to international institution [12]. Learning from past Indonesian earthquake disaster, when the international give offer to help the disaster's area then our government can't make fast decision and tend to distrust for their good deeds. Finally by our government's slow decision the victims had increased.

By making this earthquake DSS more clearly and organized then will help the government to make faster and better earthquake mitigation decision. The more complete data and variant the more accurate the decision will be created. In each of disaster's status and disaster level will have standard mitigation decision.

## 5. Conclusion

This paper as the implementation when in time none disaster research and development about disaster management can be done for support disaster management. [11, chapter 5].

Difficult for the implementation if Government do not want to make clean government, transparent and accountability.

Research for other disaster will be needed as connectivity for all disaster in one system as national disaster management.

Simulation with game approach will make more interested and increase value added for the earthquake DSS in order to educate the decision maker and citizen.

Implementation Geographical Information System on DSS will provide powerful tool for solving specific and sophisticated problems in seismology [15].